\documentclass[12pt]{article}

\usepackage[utf8]{inputenc}
\usepackage[T1]{fontenc}
\usepackage{amsmath,amssymb}
\usepackage{geometry}
\usepackage{hyperref}
\usepackage{natbib}
\usepackage{setspace}

\hypersetup{
  colorlinks=true,
  linkcolor=blue,
  citecolor=blue,
  urlcolor=blue
}

\title{\textbf{Beyond Behavior: Why AI Evaluation Needs a Cognitive Revolution}}
\author{Amir Konigsberg}
\date{}

\begin{document}

\maketitle

\begin{abstract}
In 1950, Alan Turing proposed replacing the question ``Can machines think?'' with a behavioral test: if a machine's outputs are indistinguishable from those of a thinking being, the question of whether it truly thinks can be set aside. This paper argues that Turing's move was not only a pragmatic simplification but also an epistemological commitment, a decision about what kind of evidence counts as relevant to intelligence attribution, and that this commitment has quietly constrained AI research for seven decades. We trace how Turing's behavioral epistemology became embedded in the field's evaluative infrastructure, rendering unaskable a class of questions about process, mechanism, and internal organization that cognitive psychology, neuroscience, and related disciplines learned to ask. We draw a structural parallel to the behaviorist-to-cognitivist transition in psychology: just as psychology's commitment to studying only observable behavior prevented it from asking productive questions about internal mental processes until that commitment was abandoned, AI's commitment to behavioral evaluation prevents it from distinguishing between systems that achieve identical outputs through fundamentally different computational processes, a distinction on which intelligence attribution depends. We argue that the field requires an epistemological transition comparable to the cognitive revolution: not an abandonment of behavioral evidence, but a recognition that behavioral evidence alone is insufficient for the construct claims the field wishes to make. We articulate what a post-behaviorist epistemology for AI would involve and identify the specific questions it would make askable that the field currently has no way to ask.
\end{abstract}

\section{Introduction: A Question Replaced}

In 1950, Alan Turing opened his paper ``Computing Machinery and Intelligence'' with the words: ``I propose to consider the question, `Can machines think?'\,'' Coming from the person who had formalized the concept of computation, whose codebreaking at Bletchley Park had given the Allies a crucial wartime advantage, and who was helping build one of the world's first computers, it seemed to promise a reckoning, a rigorous answer to one of the deepest questions about the nature of mind.

Yet in the very next paragraph, Turing set the question aside. Efforts to define what ``machine'' and ``think'' really mean, he argued, would spiral into precisely the kind of interminable philosophical debate that prevented progress. So rather than answer the question, he proposed to replace it with another: the imitation game. If a machine could converse so fluently that a human interrogator couldn't distinguish it from another person, Turing proposed, then we should simply treat it as thinking. The question of whether it \emph{really} thought, whether there was understanding behind the behavior, could be set aside as unanswerable and, for practical purposes, irrelevant.

This move is almost universally described as pragmatic.\footnote{The characterization of Turing's move as pragmatic is pervasive in both AI and philosophy of mind. Russell and Norvig's \emph{Artificial Intelligence: A Modern Approach} \citep{russell2010}, the field's standard textbook, presents the Turing test as a practical operationalization that sidesteps definitional debates. Dennett, in ``Can Machines Think?'' \citep{dennett1985}, explicitly defends Turing's proposal as a ``brilliantly operationalized'' way of cutting through philosophical tangles. Haugeland's \emph{Artificial Intelligence: The Very Idea} \citep{haugeland1985} frames it as an engineer's answer to a philosopher's question. Saygin, Cicekli, and Akman's survey ``Turing Test: 50 Years Later'' \citep{saygin2000} traces how the pragmatic reading became the dominant interpretation across both AI research and philosophy, noting that most subsequent discussion ``accepts Turing's basic pragmatic move and then argues about details.'' Even critics of the test, such as \citet{french1990} and \citet{harnad1991}, who argue that Turing's operationalization is insufficient, accept the characterization of the move as pragmatic before arguing that pragmatism came at a cost. The reading of Turing's proposal as an epistemological commitment rather than merely a pragmatic simplification, which is the reading this paper develops, is considerably less common. \citet{lassegue1996} and \citet{proudfoot2011} have moved in this direction, arguing that Turing's framing carried philosophical consequences beyond what the pragmatic reading acknowledges, but neither traces the consequences for AI evaluation methodology that are our focus here.} Turing made the problem tractable by translating an impossible philosophical question into a testable engineering challenge. And so he did. But what is less commonly recognized is that Turing's move was also something else: an epistemological commitment. It was a decision, embedded at the foundations of the field, about what kind of evidence would count as relevant to intelligence attribution. Behavior---and only behavior---would serve. Internal processes, mechanisms, representations, experiences: these were declared out of scope. If the outputs were right, the question was answered. If you couldn't tell the difference, there was no difference---at least none that mattered.

This paper argues that Turing's epistemological commitment has become a structural constraint on the field of artificial intelligence. It has shaped how AI evaluation benchmarks are built, how progress toward machine intelligence is measured, how competing systems are ranked, and what questions AI researchers treat as scientifically legitimate. It hasn't done any of this through explicit advocacy but rather through inheritance, with each generation of evaluation methodology being built on the assumptions of the last, and those assumptions trace back to Turing's original decision to treat behavior as sufficient evidence for intelligence attribution. The result is an evaluative infrastructure that is structurally incapable of answering the questions the field most urgently needs to ask: questions about whether high-performing systems possess the cognitive properties their benchmark scores are assumed to establish.

We are not the first to note limitations of the Turing test or of benchmark-based evaluation.\footnote{The literature on limitations of the Turing test is extensive. Searle's Chinese Room argument \citep{searle1980} contended that a system could pass the test through purely syntactic manipulation without any understanding of meaning. \citet{block1981} argued that a sufficiently large lookup table could in principle pass the test, demonstrating that behavioral indistinguishability does not entail intelligence. \citet{french1990} showed that the test fails to probe subcognitive processes that distinguish human cognition from simulation. \citet{harnad1991} argued that the test's restriction to linguistic behavior was insufficient and proposed a ``Total Turing Test'' requiring sensorimotor grounding. \citet{hayes1995} argued that the test had become a distraction from productive AI research. Dreyfus, in \emph{What Computers Can't Do} \citep{dreyfus1992}, mounted a sustained philosophical critique of the assumptions underlying AI's behavioral orientation. On the benchmark side, \citet{chollet2019} argued that existing AI benchmarks measure narrow task-specific skill rather than intelligence and proposed an alternative grounded in algorithmic information theory. \citet{raji2021} critiqued the culture of benchmark-driven evaluation, showing how it incentivizes overfitting to specific tasks at the expense of genuine capability. \citet{bowman2021} catalogued the structural problems with NLP benchmarks, including data contamination, annotation artifacts, and the gap between benchmark performance and real-world competence. Each of these critiques identifies genuine problems. What distinguishes the present paper is that it locates the source of these problems not in the design of particular tests but in the epistemological framework that governs what all such tests can tell us.} But our argument is different in kind from the usual critique. We do not argue that current benchmarks are too easy, too narrow, too contaminated, or too gameable, though they are all of these things. We argue that the \emph{epistemological framework} within which benchmarks operate forecloses a class of questions that cannot be recovered by building better benchmarks within the same framework. The problem is not the tests. The problem is the theory of evidence that governs what tests can tell us.

The parallel we draw is to the behaviorist-to-cognitivist transition in psychology.\footnote{The standard histories of this transition include Gardner's \emph{The Mind's New Science: A History of the Cognitive Revolution} \citep{gardner1985}, which traces how the behaviorist consensus unravelled across psychology, linguistics, computer science, anthropology, neuroscience, and philosophy between the mid-1950s and the early 1980s, and Baars's \emph{The Cognitive Revolution in Psychology} \citep{baars1986}, which documents the shift through interviews with its principal figures. Miller's retrospective ``The Cognitive Revolution: A Historical Perspective'' \citep{miller2003} offers a first-person account by one of its architects, dating the revolution's symbolic beginning to September 11, 1956, when he, Chomsky, and Newell and Simon presented papers at the same MIT symposium on information theory. The key primary documents of the transition are cited in the body of this paper: \citet{watson1913}, \citet{chomsky1959}, \citet{miller1960}. For a treatment that emphasizes the epistemological character of the shift, as distinct from the institutional or sociological character, see \citet{bechtel2012}, who argue that what changed was not primarily the subject matter but the standards of explanation the field was willing to accept.} For roughly half a century, from Watson's 1913 manifesto through the late 1950s, academic psychology operated under an epistemological commitment structurally identical to the one governing AI evaluation today: only observable behavior counted as scientific evidence. Internal mental states, beliefs, representations, reasoning processes, were declared unobservable and therefore outside the scope of scientific inquiry.\footnote{Watson's 1913 manifesto stated the exclusion explicitly: psychology, he argued, should be ``a purely objective experimental branch of natural science'' whose ``theoretical goal is the prediction and control of behavior,'' with introspection forming ``no essential part of its methods.'' His \emph{Behaviorism} \citep{watson1924} went further, denying the existence of ``mental existences or mental processes of any kind.'' Skinner's \emph{Science and Human Behavior} \citep{skinner1953} offered the most systematic elaboration, arguing that private events were unsuitable objects of scientific study because they were not publicly observable and could not be independently verified. Hull's \emph{Principles of Behavior} \citep{hull1943} formalized the program mathematically, constructing a stimulus-response framework that deliberately excluded reference to mental states. Skinner later drew a useful distinction, in \emph{About Behaviorism} \citep{skinner1974}, between methodological behaviorism (which denied that internal states could be accessed scientifically) and his own radical behaviorism (which acknowledged their existence but denied their explanatory value). The exclusion was not without internal dissent: Tolman's ``Cognitive Maps in Rats and Men'' \citep{tolman1948} introduced purpose and expectation into a nominally behaviorist framework, presaging the cognitive turn. But Tolman's work was the exception that proved the rule, tolerated within the paradigm precisely because it could be reinterpreted in stimulus-response terms.} This commitment made psychology rigorous by one definition of rigor. It also made it incapable of explaining the phenomena it most needed to explain. The cognitive revolution that began in the late 1950s was not primarily a change in subject matter. It was an epistemological shift: a change in what counted as legitimate evidence and legitimate questions for the scientific study of mind. It made internal processes scientifically askable again.

We argue that AI research requires an analogous shift. Not an abandonment of behavioral evidence (behavior remains indispensable) but a recognition that behavioral evidence alone cannot support the construct claims the field wishes to make: claims that benchmark performance constitutes evidence of reasoning, understanding, or intelligence, rather than evidence merely of correct outputs achieved by unknown means. The field claims to know \emph{what} a system is (i.e., it reasons, it understands) on the basis of evidence that establishes only \emph{that} it does something (it produces correct outputs). The questions that would bridge this gap, such as: Does this system reason? Does it understand? Is its performance the product of intelligence or of sophisticated pattern completion? are questions about process, not output. And the epistemological framework the field inherited from Turing has no place for them.\footnote{The practice of making construct claims on behavioral evidence is pervasive enough that it often passes without remark. \citet{bubeck2023} is perhaps the most explicit instance: the paper attributes general intelligence to a system on the basis of its performance across a range of tasks, while acknowledging that ``our approach is necessarily subjective and informal'' and that ``we do not claim that GPT-4 has genuine understanding.'' The title makes the construct claim; the caveats retract it; the framing lets both stand. OpenAI's GPT-4 technical report \citep{openai2024} describes the system as exhibiting ``human-level performance'' on professional and academic benchmarks, a performance claim that is then routinely interpreted, in the report and in subsequent coverage, as evidence of reasoning and understanding. Google DeepMind's Gemini technical report \citep{gemini2023} attributes ``reasoning capabilities'' and ``understanding'' to its system on the basis of benchmark scores. The pattern extends to domain-specific evaluations: when a language model achieves a passing score on the US Medical Licensing Examination \citep{kung2023} or the bar exam \citep{katz2024}, the finding is reported as evidence that the system ``demonstrates medical reasoning'' or ``possesses legal knowledge,'' construct claims that the behavioral evidence, however impressive, does not establish. In each case, the inferential structure is the same: a performance result (benchmark score) is treated as evidence for a construct (reasoning, understanding, intelligence) without any investigation of the process that produced the performance.}

\section{The Behaviorist Inheritance}

\subsection{What Turing Actually Did}

It is worth being precise about the nature of Turing's move, because its significance is easily underestimated. Turing did not merely propose a test for machine intelligence. He proposed a \emph{criterion}: a sufficient condition for intelligence attribution. If a system passes the imitation game, then the question of whether it thinks is answered---not deferred, not bracketed, but answered. The behavior \emph{is} the evidence, and the evidence is \emph{sufficient}.

This is a stronger claim than it initially appears. Consider what it excludes. Under Turing's criterion, two systems that produce identical outputs are, for the purposes of intelligence attribution, identical---regardless of the mechanisms that produced those outputs. A system that arrives at correct answers through compositional reasoning and a system that arrives at the same answers through memorization of a lookup table are indistinguishable. A system that generates novel solutions through genuine problem-solving and a system that retrieves pre-computed solutions from training data are equivalent. The mechanism is declared irrelevant to the attribution.\footnote{The canonical demonstration that behavioral equivalence does not entail computational equivalence is Block's lookup table argument \citep{block1981}. Block showed that, in principle, a system consisting of nothing more than a sufficiently large table mapping every possible input to an appropriate output could produce behavior indistinguishable from that of a genuinely intelligent agent, while possessing no internal structure, no reasoning, and no understanding whatsoever. The argument establishes that for any behavioral criterion of intelligence, there exists a system that satisfies the criterion through a mechanism that no one would call intelligent. The framework that makes this distinction precise is Marr's \citep{marr1982} three levels of analysis, developed within cognitive science as part of the post-behaviorist turn: the computational level (what the system computes), the algorithmic level (how it computes it), and the implementational level (what physical mechanism carries out the algorithm). Turing's criterion operates exclusively at the computational level: does the system produce the right input-output mapping? Marr's framework insists that genuine understanding of a system requires all three levels, and that two systems with identical computational-level descriptions can differ profoundly at the algorithmic level. Pylyshyn's \emph{Computation and Cognition} \citep{pylyshyn1984} develops a related argument, contending that cognitive science must distinguish between the computational structure of a process and its merely behavioral description, because the same behavior can be produced by processes with fundamentally different cognitive architectures. The passage in the text restates Block's point in contemporary terms: a system that reasons compositionally and a system that memorizes a lookup table can produce identical outputs, and no behavioral test can tell them apart.}

This exclusion was deliberate. Turing's paper anticipates and explicitly addresses the objection that behavioral evidence might be insufficient---the ``argument from consciousness,'' as he calls it, quoting the neurosurgeon Geoffrey Jefferson's 1949 Lister Oration: ``not until a machine can write a sonnet or compose a concerto because of thoughts and emotions felt, and not by the chance fall of symbols, could we agree that machine equals brain.'' Turing dismisses this by noting that we cannot verify the inner states of other humans either. ``According to this view,'' he writes, ``the only way by which one could be sure that a machine thinks is to \emph{be} the machine and to feel oneself thinking.''\footnote{Jefferson's Lister Oration, ``The Mind of Mechanical Man,'' was delivered to the Royal College of Surgeons of England on June 9, 1949, and published in the \emph{British Medical Journal} later that year \citep{jefferson1949}. Jefferson's argument was considerably more sophisticated than the excerpt Turing selected for rebuttal. Jefferson did not merely assert that machines lack consciousness. He argued that genuine intelligence requires embodied experience, that understanding is grounded in a history of sensory and emotional engagement with the world, and that a system producing correct outputs without this grounding is performing an elaborate imitation rather than thinking. The full passage continues beyond what Turing quoted: ``Not until a machine can write a sonnet or compose a concerto because of thoughts and emotions felt, and not by the chance fall of symbols, could we agree that machine equals brain---that is, not only write it but know that it had written it. No mechanism could feel (and not merely artificially signal, an easy contrivance) pleasure at its successes, grief when its valves fuse, be warmed by flattery, be made miserable by its mistakes, be charmed by sex, be angry or depressed when it cannot get what it wants.'' Jefferson's insistence on felt experience as constitutive of intelligence anticipates, by decades, the embodied cognition movement and \citeauthor{harnad1991}'s \citeyearpar{harnad1991} argument for a Total Turing Test. The exchange between Turing and Jefferson continued in a 1952 BBC Third Programme broadcast, ``Can Digital Computers Think?'', alongside the mathematician Max Newman and the philosopher Richard Braithwaite \citep[see][for a transcript and discussion]{copeland2004}. Turing's dismissal of Jefferson is itself revealing for our purposes: by arguing that we cannot verify inner states in humans either, Turing does not refute the relevance of process but declares it epistemologically inaccessible, which is exactly the move that renders process invisible within the framework he establishes.} Since we cannot access the inner states of others, he argues, we have no principled basis for demanding access to the inner states of machines. Behavior is all we have in either case.\footnote{Jefferson's Lister Oration, ``The Mind of Mechanical Man,'' was delivered to the Royal College of Surgeons of England on June 9, 1949, and published in the \emph{British Medical Journal} later that year \citep{jefferson1949}. Jefferson's argument was considerably more sophisticated than the excerpt Turing selected for rebuttal. Jefferson did not merely assert that machines lack consciousness. He argued that genuine intelligence requires embodied experience, that understanding is grounded in a history of sensory and emotional engagement with the world, and that a system producing correct outputs without this grounding is performing an elaborate imitation rather than thinking. The full passage continues beyond what Turing quoted: ``Not until a machine can write a sonnet or compose a concerto because of thoughts and emotions felt, and not by the chance fall of symbols, could we agree that machine equals brain---that is, not only write it but know that it had written it. No mechanism could feel (and not merely artificially signal, an easy contrivance) pleasure at its successes, grief when its valves fuse, be warmed by flattery, be made miserable by its mistakes, be charmed by sex, be angry or depressed when it cannot get what it wants.'' Jefferson's insistence on felt experience as constitutive of intelligence anticipates, by decades, the embodied cognition movement and \citeauthor{harnad1991}'s \citeyearpar{harnad1991} argument for a Total Turing Test. The exchange between Turing and Jefferson continued in a 1952 BBC Third Programme broadcast, ``Can Digital Computers Think?'', alongside the mathematician Max Newman and the philosopher Richard Braithwaite \citep[see][for a transcript and discussion]{copeland2004}. Turing's dismissal of Jefferson is itself revealing for our purposes: by arguing that we cannot verify inner states in humans either, Turing does not refute the relevance of process but declares it epistemologically inaccessible, which is exactly the move that renders process invisible within the framework he establishes.}

The argument has a surface plausibility that has sustained it for decades. But it treats two very different evidential situations as the same. In the human case, we have extensive convergent evidence beyond the immediate behavioral exchange---developmental history, neurological substrate, evolutionary continuity, shared embodiment---that supports the inference from behavior to inner states. When a person tells you they are in pain, you do not rely solely on their words. You know they have a nervous system that processes nociceptive signals, that they developed through a childhood in which they learned what pain means by experiencing it, that they share your evolutionary heritage as a creature for whom pain serves a survival function, that they have a body that can be injured in the way they describe. Each of these lines of evidence independently supports the inference that their verbal report reflects an inner state, and together they make the inference overwhelming. The behavioral evidence is not doing the work alone. In the machine case, this supporting evidence is absent. The behavioral evidence \emph{is} doing the work alone. The asymmetry is fundamental, and Turing's framework erases it.\footnote{The argument that we infer other minds not from behavior alone but from convergent evidence, including biological similarity, shared developmental context, and evolutionary continuity, has roots in the argument from analogy, classically associated with \citet{mill1865}: we attribute inner states to others because they resemble us in relevant respects, not merely because they behave as we do. For comprehensive treatments of the other-minds problem and the evidential structure of mental-state attribution, see \citet{avramides2001} and \citet{hyslop1995}. The use of pain as a test case connects to a long philosophical tradition. Wittgenstein's discussion of pain behavior in \emph{Philosophical Investigations} \citep{wittgenstein1953}, particularly the beetle-in-a-box thought experiment (\S293) and the extended treatment of pain language (\S244--\S271), established that the relationship between pain behavior and the inner experience of pain is neither simple nor purely behavioral: understanding what someone means when they say ``I am in pain'' depends on a shared form of life, not merely on the behavioral regularity of the utterance. The asymmetry between the human and machine cases has been noted by several authors, though typically in the context of debates about machine consciousness rather than, as here, in the context of evaluation epistemology. \citet{searle1980} argued that biological ``causal powers'' underwrite mental states in a way that syntactic manipulation cannot. \citet{nagel1974} argued that facts about conscious experience are irreducible to behavioral or functional descriptions. The contribution of the present passage is to reframe this asymmetry as an evidential problem: not a metaphysical claim about what machines lack, but an epistemological claim about the difference in the evidence available for attributing inner states in the two cases.}

\subsection{From Test to Infrastructure}

Turing's criterion might have remained a thought experiment. Instead, it became the template for an evaluative infrastructure. The specific form of the imitation game---pitting machine performance against human performance on defined tasks---was operationalized across decades of AI research in the form of benchmarks: various tests on which systems are evaluated by their output correctness, ranked by their scores on these tests, and interpreted through the lens of human-comparable performance.

The evaluative framework that declares GPT-4 intelligent is, in its logical structure, the same one that measured whether Deep Blue could beat a grandmaster. What expanded was the scope of the claim. What didn't expand was the evidence. Early AI evaluation was task-specific: chess programs were measured by their play strength, theorem provers like Newell and Simon's Logic Theory Machine by which theorems they could prove, expert systems by their diagnostic accuracy. Each evaluation was confined to a single, well-defined domain, and the inference from performance to intelligence was hedged: no one claimed that Deep Blue ``understood'' chess. Performance was treated as evidence of task-specific capability, not as evidence of cognitive properties like reasoning or understanding.\footnote{Newell and Simon's Logic Theory Machine \citep{newell1956} proved 38 of the first 52 theorems in Chapter 2 of Russell and Whitehead's \emph{Principia Mathematica}, and was evaluated strictly by which theorems it could and could not prove. Their subsequent General Problem Solver \citep{newell1963} was assessed by the range of puzzles and logical problems it could solve. In neither case was the system's performance interpreted as evidence of general intelligence or understanding; the evaluation was explicitly task-specific. Deep Blue's 1997 defeat of Garry Kasparov generated widespread commentary, but IBM's own researchers were careful to characterize the system as a chess-playing machine, not an intelligent agent. \citet{campbell2002} describe the system's architecture in entirely engineering terms: search depth, evaluation functions, special-purpose hardware. Kasparov himself drew the distinction sharply, noting that the machine could calculate but did not understand what it was doing \citep{kasparov2017}. On the expert systems side, MYCIN \citep{shortliffe1976} was evaluated by whether its antibiotic recommendations matched those of Stanford's infectious disease specialists, and it performed comparably, but neither its designers nor its evaluators claimed it possessed medical understanding. The hedging of intelligence claims in this era was deliberate and principled: researchers understood that task-specific behavioral performance, however impressive, was not sufficient evidence for cognitive attributions. What changed with LLMs was not the evidential framework but the ambition of the claims it was used to support.}

The arrival of large language models changed this. Because LLMs operate in natural language, the medium through which virtually all cognitive tasks can be expressed, a single system can now be tested on mathematical reasoning, reading comprehension, scientific knowledge, coding, common-sense inference, legal analysis, and ethical judgment. Each of these domains was previously the province of a dedicated system with its own specialized evaluation: a chess engine was tested on chess, a medical expert system on diagnosis, a theorem prover on proofs. No single system was ever evaluated across all of them simultaneously, and no researcher would have aggregated a chess rating, a diagnostic accuracy score, and a proof count into a composite measure of intelligence. LLMs made this aggregation possible because they were able to produce outputs across every domain using the same architecture and the same medium. The temptation to interpret broad benchmark performance as evidence of general intelligence follows naturally from this convergence, and the behavioral framework the field inherited offers no principled basis for resisting it.

This interpretive temptation is a direct consequence of Turing's epistemological commitment. If behavior is sufficient for intelligence attribution, and if the behavior now spans virtually every cognitive domain, then the inference to general intelligence follows naturally. The benchmarks were never designed to support this inference. They were designed to measure task-specific performance. But the epistemological framework treats task-specific performance and the possession of cognitive properties like intelligence as the same thing. The psychologist Edwin Boring captured this conflation in 1923 when he declared that ``intelligence is what the tests test,'' a formulation that was meant as a pragmatic concession but that AI evaluation has reproduced as an unexamined assumption: reasoning is what reasoning benchmarks measure, understanding is what understanding benchmarks measure.\footnote{Boring's formulation \citep{boring1923} was rooted in the broader philosophical tradition of operationalism, introduced by the physicist \citet{bridgman1927}, which held that a scientific concept is nothing more than the set of operations used to measure it: length is what rulers measure, temperature is what thermometers measure, intelligence is what intelligence tests measure. \citet{cronbach1955} developed the construct validity framework in explicit response to this operationalist collapse, arguing that psychological constructs must be defined independently of any single measurement instrument and that the validity of a test in measuring a construct is an empirical question, not a definitional one. In AI evaluation, the operationalist move has been reproduced almost exactly, as the body text notes: MMLU is described as measuring ``understanding,'' so a high score is treated as evidence of understanding; GSM8K is described as measuring ``reasoning,'' so a high score is treated as evidence of reasoning. \citet{mitchell2023} identify this pattern explicitly, arguing that the AI field routinely conflates task performance with cognitive capacities and that benchmark labels function as unexamined theoretical commitments disguised as neutral descriptions. The structural cause is the behavioral framework itself, which makes the conflation invisible by treating output as the only admissible evidence for the properties outputs are claimed to indicate.}

\subsection{The Invisible Constraint}

A framework's blind spots are harder to see than its errors. Under a strictly behavioral epistemology, certain questions cannot be formulated as scientific questions.

Consider the question: ``Does this system arrive at its answer through reasoning or through memorization?'' Under Turing's framework, this question has no methodological traction. If the output is correct, the process is irrelevant. The question is not rejected on empirical grounds, it is rejected on epistemological grounds. It asks about something (internal process) that the framework has declared outside the scope of evidence.

Or consider: ``Does this system understand the problem it has solved, or has it learned to produce the output that is statistically associated with this type of input?'' Again, the framework offers no way to investigate this. Understanding, if it means anything beyond correct output, refers to something about the system's internal relationship to the problem, and internal relationships are precisely what the behavioral framework excludes.

These are not marginal questions. They are the questions that would need to be answered in order to determine whether a system possesses the cognitive properties that benchmark performance is taken to indicate. The field treats benchmark scores as evidence of reasoning, understanding, and intelligence. But the epistemological framework on which the benchmarks are built has no resources for investigating whether those attributions are warranted, because the framework was designed, from the outset, to make the attributions unnecessary.

\section{The Parallel: Psychology's Escape from Behaviorism}

\subsection{The Behaviorist Commitment}

The parallel between AI evaluation and psychological behaviorism is structural, not just analogical. Both involve a deliberate epistemological restriction, a decision to treat only observable behavior as legitimate scientific evidence, motivated by the desire to make an otherwise intractable domain scientifically rigorous.

John B.\ Watson's 1913 paper ``Psychology as the Behaviorist Views It'' made this commitment explicitly clear. Psychology, Watson argued, should concern itself exclusively with observable behavior. Mental states, consciousness, introspection, these were unscientific, unverifiable, and unproductive. For Watson, and for the behaviorist tradition that followed through Skinner and Hull, the commitment to behavioral evidence was synonymous with scientific rigor,\footnote{Watson \citep{watson1913} is the founding document of methodological behaviorism. The paper appeared in \emph{Psychological Review} and established the programmatic commitments that would govern American experimental psychology for roughly four decades. For analysis of the epistemological (as opposed to merely methodological) character of Watson's commitment, see \citet{smith1986}, who traces the deep connections between behaviorism and logical positivism, showing that both movements shared the conviction that scientific legitimacy required restricting evidence to the publicly observable. Zuriff's \emph{Behaviorism: A Conceptual Reconstruction} \citep{zuriff1985} provides the most systematic analysis of what behaviorism actually committed to and what it excluded, distinguishing between its metaphysical claims (about what exists), its methodological claims (about what can be studied), and its epistemological claims (about what counts as evidence). It is the epistemological claim that matters for our parallel: both behaviorist psychology and Turing-framework AI evaluation restrict admissible evidence to observable behavior, and both do so for the same reason, that the alternative (admitting unobservable internal states as evidence) seems to threaten scientific rigor. The parallel has been noted in passing by several authors. \citet{buckner2024} observes that debates about LLM cognition recapitulate debates from the behaviorist era about whether internal representations can be scientifically posited. \citet{shevlin2021} argues that AI research would benefit from engaging with the philosophy of psychology's long experience in navigating between behaviorism and cognitivism. The contribution of the present paper is to develop the parallel systematically rather than noting it in passing, and to show that the epistemological restriction produces the same specific consequences in both domains: an inability to distinguish between systems that achieve identical outputs through fundamentally different processes.} and it produced genuine achievements such as precise, replicable experimental paradigms and lawful relationships between stimuli and responses. It also grounded psychology in observable data rather than subjective report. By the standards it set for itself, it succeeded.

But the restriction also produced systematic blind spots. Behaviorism could describe the relationship between a stimulus and a response but could not explain how the organism got from one to the other. It could predict behavior in controlled settings but could not account for the flexibility, generativity, and context-sensitivity of human cognition in natural settings. It could not explain language acquisition, problem-solving, planning, or any cognitive process that required positing internal representations or computational structures.\footnote{Each of the explanatory failures named in the text has a specific history. Chomsky's review of Skinner's \emph{Verbal Behavior} \citep{chomsky1959} remains the most celebrated: Chomsky demonstrated that stimulus-response theory could not account for the productivity of language, the fact that children routinely produce and understand sentences they have never encountered, because productivity requires positing internal generative rules that the behaviorist framework excluded by fiat. Lashley's ``The Problem of Serial Order in Behavior'' \citep{lashley1951} posed the planning problem: complex sequential behavior, from playing a musical phrase to constructing a sentence, requires hierarchical organization that cannot be explained as a chain of stimulus-response associations, because the later elements of the sequence are planned before the earlier elements are executed. The flexibility and context-sensitivity of human cognition was demonstrated experimentally by studies like \citet{bruner1956}, which showed that human categorization involves active hypothesis testing, a process that requires internal representations of candidate categories and strategies for revising them. Newell and Simon's work on problem-solving \citep{newell1956,newell1972} showed that even simple puzzle-solving involves search through a space of possible states, a process that is invisible to behavioral observation and can only be characterized by positing internal representations of goals, operators, and intermediate states. Miller's ``The Magical Number Seven'' \citep{miller1956} demonstrated that the capacity limits of human information processing could not be explained without positing internal memory structures. Taken together, these findings did not merely suggest that behaviorism was incomplete. They demonstrated that the phenomena psychology most needed to explain, the ones that make human cognition distinctively flexible and generative, were precisely the phenomena the epistemological restriction made unexplainable.}

\subsection{The Cognitive Revolution}

The cognitive revolution that began in the late 1950s, marked conventionally by Chomsky's 1959 review of Skinner's \emph{Verbal Behavior}, Newell and Simon's work on the General Problem Solver, and Miller, Galanter, and Pribram's \emph{Plans and the Structure of Behavior} (1960), was not primarily a change in subject matter. Psychologists did not suddenly become interested in thinking. They had always been interested in thinking. What changed was the epistemological framework: the rules governing what counted as legitimate evidence and legitimate explanation.

The key shift was the acceptance of internal representations and computational processes as scientifically legitimate constructs, things that could be posited, theorized about, and indirectly tested through their behavioral consequences, even though they could not be directly observed. This was not a return to introspection. It was the development of a new evidential standard: internal processes could be inferred from behavioral evidence, but only through converging lines of evidence that constrained the inference---reaction times, error patterns, priming effects, dissociations, transfer profiles. The inference from behavior to process was indirect but disciplined. Consider what this looked like in practice. When \citet{shepard1971}\footnote{Shepard and Metzler's study, published in \emph{Science} in 1971, presented participants with pairs of line drawings depicting three-dimensional block structures. In each pair, the two structures were either identical but rotated to different orientations, or mirror images of each other. Participants had to judge whether the two structures were the same object or different objects. The key finding was that response time was a linear function of the angular difference between the two orientations, with a slope of roughly one second per 60 degrees of rotation, regardless of whether the rotation was in the picture plane or in depth. This linearity was the evidence. If participants had been comparing feature lists or matching templates, there would be no reason for the angular difference to matter, and certainly no reason for the relationship to be linear. The linearity implied that participants were performing an internal operation that preserved the geometric structure of physical rotation: they were rotating a mental representation through the intervening angle, and this operation took time proportional to the distance traversed. Shepard later extended this work into a broader program of research on the second-order isomorphism between mental representations and the physical structures they represent \citep{shepard1970,shepard1984}, arguing that internal representations share not the surface properties of external objects but their relational structure. \citet{cooper1973} provided additional evidence by showing that people could be primed to rotate in a specific direction and that partially rotated stimuli produced intermediate response times, consistent with an incremental rotation process. The study became a paradigm case for the cognitive revolution's central methodological claim: that the structure of an internal process can be inferred, with precision, from the pattern of its behavioral effects.} asked people to judge whether two images depicted the same three-dimensional object rotated to different orientations, they found that response time increased linearly with the angle of rotation, as though participants were mentally rotating one object to match the other. No one could observe the rotation happening. But the linear relationship between angle and response time constrained the inference so tightly that the internal process had to have a structure that mirrored physical rotation. A purely behavioral account, one that recorded only whether the response was correct, would have missed this entirely. The reaction time data revealed the process, and the process was the finding.\footnote{The philosophical justification for treating internal representations as scientifically legitimate was developed most explicitly by \citet{fodor1975}, who argued that cognitive processes are computational operations over internal representations and that this claim is empirically testable through its behavioral predictions. Pylyshyn's \emph{Computation and Cognition} \citep{pylyshyn1984} extended this argument, contending that the computational level of description is both scientifically indispensable and distinct from the behavioral level. The specific methodological innovations that made internal processes inferable from behavioral evidence have their own history. The use of reaction time as a window into internal processing traces back to Donders' subtraction method \citep{donders1969}, which was rediscovered and refined by cognitive psychologists in the 1960s. \citet{sternberg1969} developed the additive factors method, showing that if two experimental manipulations affect reaction time independently (additively), they must influence different processing stages, allowing the internal architecture of a task to be inferred from the pattern of reaction time effects. Posner's \emph{Chronometric Explorations of Mind} \citep{posner1978} provided the most comprehensive demonstration that reaction time methods could reveal the temporal structure of internal processing with millisecond precision. Priming effects, in which exposure to one stimulus influences the processing of a subsequent stimulus, were developed as a tool for probing the organization of internal representations by \citet{meyer1971}. Double dissociations, in which two patients show complementary patterns of preserved and impaired abilities, were established as evidence for the functional independence of cognitive subsystems by \citet{teuber1955} and formalized by \citet{shallice1988}. The \citet{shepard1971} finding discussed in the text is a canonical example of how these methods converge: the reaction time data not only reveals that an internal process is occurring but constrains its structure so tightly that the process must have a specific geometric character.}

The crucial point is that the cognitive revolution did not reject behavioral evidence. It rejected the sufficiency of behavioral evidence. Behavior remained indispensable; you could not study cognition without it. But behavior alone could not answer the questions cognitive science needed to ask. The questions were about process, mechanism, and representation. Answering them required supplementing behavioral data with theoretical models of internal computation, and testing those models against behavioral predictions that different internal processes would generate.\footnote{Fodor's \emph{Psychological Explanation} \citep{fodor1968} articulated the core argument: the same behavioral output can be produced by different internal mechanisms, so explaining behavior requires specifying which mechanism produced it. This is the underdetermination problem the present paper identifies in AI evaluation. Neisser's \emph{Cognitive Psychology} \citep{neisser1967}, the book that named the field, was built around the methodology the passage describes: positing internal processes, building computational models, and testing them against behavioral predictions. The imagery debate demonstrated this methodology vividly: \citet{kosslyn1980} and \citet{pylyshyn1981} proposed structurally different models of mental representation that were both consistent with existing data but generated different predictions about specific experimental conditions. \citet{anderson1978} raised the deeper question of whether behavioral evidence alone could ever distinguish between competing models of internal representation, arguing that in some cases structurally different models are behaviorally indistinguishable. This is the AI evaluation problem in microcosm.}

\subsection{What the Change Made Clear}

The cognitive revolution didn't just introduce new themes for the field of psychology to explore. It also showed that the epistemological restriction of behaviorism had been preventing the field from attending to phenomena that were in plain sight. To see this, consider Chomsky's argument about language acquisition. Children acquire language with a speed and generativity that simply cannot be explained by the stimulus-response learning mechanisms that behaviorism allowed for.\footnote{The claim that behaviorist learning theory cannot account for language acquisition is the central argument of Chomsky's review of Skinner's \emph{Verbal Behavior} \citep{chomsky1959}. Chomsky identified several specific failures: children produce and understand sentences they have never been exposed to (the productivity problem), they do so from early childhood with remarkably little explicit instruction (the poverty of the stimulus problem), and they converge on the same grammatical structures despite wide variation in the language they hear (the invariance problem). Each of these phenomena requires positing internal structure, specifically a generative grammar, that the child brings to the task and that the stimulus-response framework had no way to represent. For a comprehensive treatment of the poverty of the stimulus argument and its subsequent development, see \citet{laurence2001}. For an account of the broader impact of Chomsky's review on the decline of behaviorism, see \citet{gardner1985}, Chapter~7.} Children produce sentences they have never heard before, some of which no one has ever said before, and they apply grammatical rules to words they have never been exposed to. In Jean Berko's classic demonstration (1958), children shown a picture of a novel creature called a ``wug'' were then shown two of these, which they referred to as ``wugs,'' successfully applying the English plural rule to a word they had never heard and that didn't even exist. They make errors (such as overregularization: ``I goed,'' ``two mouses'') that demonstrate they are applying rules rather than just reproducing from memory. Importantly, these were not new phenomena for behaviorist researchers, yet their epistemological framework had no way to explain them, because explanation required positing internal structures, grammars, rules, representations, that the framework excluded on principle.\footnote{Berko's ``Wug Test'' \citep{berko1958} remains one of the most elegant demonstrations in developmental psycholinguistics. By using nonsense words, the study eliminated the possibility that children were reproducing forms they had memorized from adult speech: no child had ever heard the word ``wugs,'' so producing it required applying a rule. The study tested not only plurals but also possessives, past tenses, and progressive forms, and children as young as four performed reliably, demonstrating productive rule application across multiple morphological domains. Overregularization errors like ``goed'' and ``mouses,'' documented extensively by \citet{marcus1992}, provide converging evidence: these forms cannot be imitations of adult speech (adults don't say ``goed''), so they must be the product of a rule applied where it shouldn't be, which means the child has a rule. The broader point for our argument is that behaviorism's epistemological framework could observe these phenomena, it could record the child saying ``wugs'' and ``goed,'' but it could not explain them, because the explanation requires reference to internal rules that are not observable in the behavioral data. The framework could describe the output but not the process that produced it, which is precisely the limitation we identify in AI evaluation.}

AI evaluation faces the same problem. Current benchmarks, MMLU for broad knowledge, GSM8K for mathematical reasoning, HumanEval for code generation, ARC for abstract reasoning, can show that a system produces correct, high-quality, or low-error-rate outputs.\footnote{The benchmarks named in the text represent the primary evaluation instruments used to assess and compare current frontier language models. MMLU \citep{hendrycks2021} tests performance across 57 academic subjects, from abstract algebra to virology, and is routinely cited as evidence of broad knowledge and reasoning. As of early 2025, leading models exceed 90\% accuracy. GSM8K \citep{cobbe2021} tests grade-school mathematical word problems requiring multi-step reasoning; frontier models now approach near-perfect scores on the original test set, prompting the development of harder variants. HumanEval \citep{chen2021} measures code generation by testing whether model-produced programs pass unit tests; it has been supplemented by HumanEval+ and other extensions as models saturated the original. ARC-AGI \citep{chollet2019} was explicitly designed to test general fluid intelligence through abstract pattern-completion tasks that resist memorization; it remains the most ambitious attempt to measure something beyond task-specific skill, though even Chollet's framework remains behavioral in the sense this paper identifies: it evaluates the output (whether the pattern is completed correctly), not the process by which the system arrived at it. GPQA \citep{rein2024} tests graduate-level scientific reasoning with questions designed to be resistant to search engine lookup. In each case, the benchmark measures output quality. None provides evidence about the computational process that produced the output, which is the gap this passage identifies.} But they cannot observe, and the existing epistemological framework does not permit asking, whether the system produces those outputs by combining learned principles in new configurations (the way a person who understands arithmetic can solve problems they have never seen before), by retrieving stored patterns from training data (such that encountering a sufficiently similar problem during training is what drives the correct answer, not any general capacity), by exploiting statistical regularities in the way benchmark questions are phrased, or through some process that does not map onto any of these categories. These are not questions about consciousness or subjective experience. They are questions about computational mechanism, precisely the kind of questions that the cognitive revolution made legitimate in the study of human minds, and that remain illegitimate, by epistemological inheritance, in the study of artificial ones.

\section{The Epistemological Ceiling}

\subsection{What Behavioral Evaluation Cannot Tell Us}

The argument of the previous sections can be stated precisely. There exists a class of questions that are essential to intelligence attribution and that behavioral evaluation is, in principle, incapable of answering. These questions concern the relationship between a system's outputs and the processes that produce them.

We can formalize this. Let $O$ denote the set of possible outputs a system can produce on a given benchmark. Let $M$ denote the set of possible computational mechanisms that could produce those outputs. For any output $o \in O$, there exists a set $M(o) \subseteq M$ of mechanisms capable of producing $o$. Behavioral evaluation observes $o$. It does not observe which element of $M(o)$ produced it.

If all elements of $M(o)$ warranted the same intelligence attribution, this would not matter. But they do not. $M(o)$ contains mechanisms that involve genuine compositional reasoning (breaking a problem into parts, applying specific principles to each, and combining the results), mechanisms that involve memorization (matching the input to a sufficiently similar example encountered during training and reproducing the associated output), mechanisms that involve shortcut exploitation (keying on superficial features of the input, such as the format of a multiple-choice question or the statistical association between certain words and certain answer types, rather than engaging with the problem's structure), and mechanisms that involve processes we do not yet have adequate descriptions for (hybrid strategies, partial reasoning combined with pattern completion, or computations that don't cleanly map onto any human cognitive category). A system that solves a mathematical problem by decomposing it into subproblems and applying algebraic rules to each is doing something fundamentally different from a system that produces the same correct answer because it encountered a nearly identical problem during training. The first has a capacity that generalizes; the second has a match that may not. The question of which mechanism produced the output is therefore the question on which intelligence attribution depends. And it is the question that behavioral evaluation, because it observes only inputs and outputs and nothing in between, cannot answer.

This is not a contingent limitation that could be overcome by building better benchmarks. It is a structural property of the behavioral framework. Any evaluation that observes only inputs and outputs and infers cognitive properties from the mapping between them faces the same underdetermination. The mapping is many-to-one: multiple mechanisms produce identical outputs. And the information lost in this many-to-one mapping is precisely the information needed for intelligence attribution.

\subsection{The Ceiling in Practice}

This underdetermination is not hypothetical.

\paragraph{Did it reason, or did it remember?} A leading language model scores 95\% on a benchmark of grade-school math problems. Researchers then add a single irrelevant sentence to each problem, something like ``There were also 15 books on the shelf'' inserted into a problem about calculating the cost of apples. The math is unchanged. The correct answer is unchanged. Performance drops by over 20 percentage points. Systems that achieve high scores on mathematical reasoning benchmarks have been shown, repeatedly, to fail under perturbations that preserve logical structure but alter surface features \citep{shi2023,nezhurina2024}. This finding is consistent with memorization-based strategies and inconsistent with genuine reasoning. A person who understands arithmetic doesn't get confused by an irrelevant sentence; a system that has learned to match problems to similar ones it encountered during training does, because the irrelevant sentence makes the problem look different from anything it has seen. But the original benchmark scores were identical for systems that fail under perturbation and systems that do not. The behavioral evaluation could not distinguish them. The distinction became visible only when researchers moved beyond the behavioral framework and began testing robustness, probing for process-level evidence, in other words, when they violated the epistemological restriction the framework imposes.\footnote{Shi et al.\ \citep{shi2023} systematically tested the effect of adding irrelevant context to GSM8K problems, finding that performance of GPT-3.5 and other models dropped substantially, in some conditions by more than 20 percentage points, when a sentence containing irrelevant numerical information was inserted into an otherwise unchanged problem. The logical structure was identical; only the surface features changed. A genuine reasoner should be unaffected by irrelevant information; that performance degrades indicates sensitivity to surface statistics rather than engagement with logical structure. \citet{nezhurina2024} demonstrated an even starker case: frontier models that achieved near-perfect scores on standard reasoning benchmarks failed elementary problems when presented in slightly unfamiliar phrasings, producing confidently wrong answers to questions a child could handle. The broader literature on adversarial robustness in NLP tells the same story at scale: \citet{ribeiro2020} developed CheckList, a systematic testing methodology that revealed that models performing at benchmark ceilings failed under simple perturbations such as negation, name substitution, and paraphrase. \citet{wang2021} constructed Adversarial GLUE, showing that models scoring above 90\% on standard NLU benchmarks dropped to near chance under adversarial rephrasing. What matters for our argument is not merely that these failures occur but what their existence demonstrates about the original benchmark scores: the scores were uninformative about the process, because the same score was achieved by systems that were reasoning and systems that were not, and nothing in the behavioral evaluation could tell the difference.}

\paragraph{The faithfulness problem.} Chain-of-thought prompting was introduced partly to make the reasoning process visible---to move beyond pure behavioral evaluation by examining the steps by which the system arrives at its answer \citep{wei2022}. But subsequent research has shown that these reasoning traces are frequently unfaithful: the stated steps do not reliably reflect the actual computational process that produced the output \citep{turpin2024,lanham2023}. The system generates a plausible-looking reasoning trace alongside a correct answer, but the trace and the answer may be produced by independent processes. This is a particularly revealing case, because it shows that even an attempt to examine process within the behavioral framework---by asking the system to display its process---fails, because the displayed process is itself a behavioral output subject to the same underdetermination.\footnote{Chain-of-thought prompting was introduced by \citet{wei2022}, who showed that prefacing a question with step-by-step worked examples caused language models to produce intermediate reasoning steps before their final answer, and that this substantially improved performance on arithmetic, common-sense, and symbolic reasoning tasks. The technique was motivated in part by the intuition that making the reasoning process visible would provide evidence about how the system was arriving at its answers, not just whether the answers were correct. It appeared to offer precisely the kind of process-level evidence that the behavioral framework lacked. Subsequent research has undermined this promise. \citet{turpin2024} demonstrated that chain-of-thought traces are systematically unfaithful: when models were given biased cues (such as a suggestion from a professor that a particular multiple-choice answer was correct), the cues influenced the final answer but did not appear in the reasoning trace. The model's stated reasoning presented a clean logical path to a conclusion that was actually driven by a biased cue it never mentioned. \citet{lanham2023} found similar patterns through a different methodology, showing that truncating or corrupting the chain of thought often had little effect on the final answer, suggesting that the reasoning trace and the answer are produced by partially independent processes rather than the trace driving the answer. OpenAI's own evaluation of its reasoning models \citep{openai2025} acknowledges that chains of thought ``are not complete or fully transparent accounts of their internal computations.'' \citet{wang2023} showed that sampling multiple chains of thought from the same model and taking the majority answer (self-consistency) improves accuracy, which is revealing: if the chains of thought faithfully represented the model's reasoning, there would be no reason for different samples to disagree. The deeper point for our argument is that chain-of-thought is itself a behavioral output. Asking a model to show its work is asking it to generate text that looks like reasoning, and a system optimized to produce text that looks like reasoning will produce text that looks like reasoning regardless of whether it is reasoning. The displayed process is subject to the same underdetermination as the answer it accompanies.}

\paragraph{The RLHF circularity.} Reinforcement learning from human feedback trains systems to produce outputs that human evaluators prefer \citep{christiano2017,ouyang2022}. Human evaluators prefer responses that appear thoughtful, coherent, and intelligent. The training process therefore optimizes for the appearance of intelligence as judged by human behavior. Under a behavioral epistemology, this appearance \emph{is} intelligence. The circularity is complete: the system is trained to produce behavior that satisfies the criterion that behavior is sufficient, and the evaluation framework that declares the criterion satisfied is the same framework that defined it. No external check is possible from within the framework.

\subsection{What We'd Need to Know, and Can't}

AI systems are now being used to assist medical diagnoses, build legal cases, evaluate job candidates, tutor children, summarize scientific research, and inform policy decisions. Each of these applications carries an implicit assumption about the cognitive properties of the system performing the task. Those using a diagnostic assistant assume it is reasoning about symptoms rather than pattern-matching against training cases, which may not include the patient's specific condition. Those using a legal tool assume it understands the argument it is constructing, rather than producing text that has the look and form of legal reasoning. Those relying on an AI tutor assume it grasps the concept it is explaining, rather than generating a statistically likely sequence of pedagogical-sounding sentences.

Before trusting a system in any of these roles, we'd want to know certain things. Such as whether it breaks problems into components and applies relevant principles to each, or whether it matches a given input to similar cases it encountered during training and reproduces what worked there. This matters concretely: a diagnostic tool that works through what the symptoms mean in combination can handle a patient whose combination of symptoms doesn't appear in the medical literature, while one that matches patterns may miss the diagnosis entirely because it has never seen that particular combination. We'd want to know whether the way it handles problems it hasn't encountered before reflects a genuine capacity to generalize, or merely reflects its training data being dense enough that most problems it encounters are close to something it has already seen, because a legal tool that generalizes can construct an argument for a novel kind of case, while one that interpolates will produce plausible-sounding nonsense when the case is unlike anything it was trained on. We'd want to know how it behaves when conditions shift, because a tutor that recognizes when it has reached the boundary of what it knows can say so, while one that doesn't will be confidently wrong without warning. And we'd want to know whether what it has learned transfers to structurally related problems in new domains, because a system that transfers has acquired something like a capability, while a system that doesn't has memorized something like a script.

The behavioral framework cannot provide any of this knowledge. It can tell us that the system produces correct outputs on a benchmark. It cannot tell us which of these very different processes produced them. And the distinction matters, practically and immediately, because a system deployed in a hospital, a courtroom, or a classroom is being trusted with a role that presupposes cognitive properties the evaluation never established.

\section{Toward a Post-Behaviorist Epistemology}

\subsection{What the Transition Requires}

The cognitive revolution in psychology did not reject behavioral evidence. It supplemented it with converging evidence from multiple sources, such as reaction times, error patterns, neuroimaging, developmental trajectories, lesion studies, computational modeling, that made it possible to draw disciplined conclusions about what was happening inside the mind. Behavior remained the foundation. But the sufficiency claim was abandoned: behavior was necessary but not sufficient for claims about cognitive processes and representations.

An analogous transition in AI evaluation would involve four commitments.

The first is the explicit recognition that benchmark performance is evidence of performance, not evidence of cognitive properties. The sentence ``GPT-4 achieves 86\% on MMLU'' is a performance claim. The sentence ``GPT-4 demonstrates broad language understanding'' is a construct claim. These are different claims, requiring different evidence. The field (and increasingly also the general public using these systems) currently treats them as interchangeable. A post-behaviorist epistemology would insist on the distinction and demand separate evidential support for each.

The second is the development of methods for generating process-level evidence, evidence about the computational mechanisms by which systems arrive at their outputs. The most promising current direction is what researchers call mechanistic interpretability: opening up a model's internals to examine which circuits activate, which representations form, and how information flows from input to output. But it is not the only avenue. Systematically altering inputs to see what the system is actually sensitive to (the kind of perturbation testing described in Section~4.2), testing whether what a system learned in one domain carries over to another, tracking how a system's capabilities change across the course of training, and selectively disabling components to see what breaks, all provide indirect evidence about internal processes. The key shift is treating this evidence as relevant to intelligence attribution, rather than as merely interesting technical detail.

The third is the construction of theoretical frameworks that connect internal processes to behavioral predictions, the AI equivalent of cognitive models. If a system reasons compositionally, what behavioral signature should this produce? If it memorizes, what different signature? If it exploits shortcuts, what different signature again? These questions require theories that map mechanisms to observables, and they require benchmark designs that can discriminate between the predictions of competing theories. For example, if you suspect that a system might be memorizing rather than reasoning, you would design a test that includes problems with familiar structure but novel content alongside problems with novel structure but familiar content. A compositional reasoner should handle the first kind easily and struggle with the second. A memorizer should show the opposite pattern. The diagnosis comes not from whether the system gets the right answer but from which conditions cause it to fail, because different mechanisms fail differently. This is precisely the methodology the cognitive revolution introduced to psychology, and it is precisely what is missing from AI evaluation.

The fourth is developmental evaluation. Rather than evaluating systems at a single point in time, researchers could examine how capabilities develop across training. Whether a system shows phase transitions, U-shaped learning curves, and error patterns characteristic of genuine learning, or whether performance improves monotonically in a way consistent with memorization of an expanding training set, is evidence about process that single-point behavioral evaluation discards.

\subsection{The Limits of the Analogy}

This analogy has limits, which need to be made clear.

The entities under study in psychology, before, during, and after behaviorism, are human beings, recognized as having internal processes. The contestable question was whether those processes could be studied scientifically. The cognitive revolution answered yes. Yet when it comes to AI, it isn't known whether the entities under study have internal processes of the relevant kind, whether, that is, the computational operations running within a neural network constitute ``reasoning'' or ``representation'' in anything more than a metaphorical sense. The cognitive revolution in psychology was epistemological: it changed what counted as evidence. The corresponding shift in AI would need to be both epistemological and ontological: it would need to develop methods for generating process-level evidence and frameworks for determining what kind of processes warrant cognitive description.

This additional challenge does not weaken the argument for the transition to a post-behaviorist epistemology. If we do not know whether AI systems have internal processes that warrant cognitive description, then we certainly cannot infer from behavioral evidence alone that they do. The behavioral framework assumes the answer to the ontological question (outputs suffice, so the question doesn't matter) rather than investigating it. A post-behaviorist framework would treat it as an open empirical question, which is what it is.

A second point is that psychology's cognitive revolution was enabled by the development of information-processing models that could be tested against behavioral data. Cognitive scientists could posit internal representations and derive behavioral predictions that discriminated between competing models. Sperling (1960), for instance, hypothesized that visual information is held briefly in a high-capacity sensory store that decays within a fraction of a second. This posit generated a specific prediction: if you flash a grid of letters and then cue participants to report just one row, they should succeed when the cue comes immediately but fail as the delay increases, because the store is decaying. That is exactly what happened, and the decay curve's shape discriminated Sperling's model from alternatives that posited no such store.\footnote{Sperling's experiment \citep{sperling1960} is a paradigm case of the cognitive revolution's methodology. Participants were shown a grid of letters (typically three rows of four) for 50 milliseconds, then asked to report what they had seen. In the whole-report condition, they could typically recall only four or five of the twelve letters. This seemed to suggest a severe capacity limit. But Sperling hypothesized that participants had actually seen all the letters and that the information was decaying from a high-capacity sensory store during the time it took to report. To test this, he introduced a partial-report condition: a tone sounded after the display, indicating which row to report. When the tone came immediately, participants could report nearly all the letters in the cued row, implying access to the full grid. As the delay between display and tone increased, performance declined toward whole-report levels, tracing a decay function that characterized the temporal properties of the store. The critical point is methodological: an internal process (rapid sensory decay) that could not be directly observed was inferred from a pattern of behavioral results (the interaction between delay and report accuracy) that different theories of the internal process predicted differently. A theory with no sensory store predicted no partial-report advantage at any delay. A theory with a permanent store predicted a partial-report advantage at all delays. Only a theory with a rapidly decaying store predicted the specific pattern Sperling observed. The behavior constrained the process. This is the evidential logic that AI evaluation currently lacks.} In AI, the systems themselves are available for direct inspection in ways that human brains are not; we can examine weights, activations, attention patterns, and computational graphs. This is an advantage that the cognitive revolution in psychology did not have. The tools for generating process-level evidence already exist, at least in embryonic form. What is missing is the epistemological framework that treats this evidence as relevant to the field's central claims.

\section{From Turing-Thinking to Construct-Thinking}

\subsection{Three Questions Treated as One}

Turing's 1950 paper was called ``Computing Machinery and Intelligence,'' but its central question was about thinking, and it used the terms ``intelligence'' and ``thinking'' interchangeably. We believe this was not accidental. The imitation game was designed to make the differences between intelligence, thinking, and understanding irrelevant. In ordinary language, these words pick out somewhat different (though closely related) properties. Intelligence relates to problem-solving, adapting, and achieving goals, what a system can do. Thinking suggests something happening inside, a process of reasoning or deliberation, how a system arrives at answers. Understanding implies grasping what things mean, not just manipulating the symbols that represent them.

Turing's test reduced all three into a single behavioral criterion: does the output convince? This collapse was productive for engineering. It gave researchers a clear target. But it was destructive for science, because it erased distinctions that a science of machine intelligence would need to make. The question of whether a system is intelligent (can it solve the problem?) is different from the question of whether it thinks (does it arrive at the solution through a process that warrants being called reasoning?) which is different from the question of whether it understands (does it grasp what the problem means?). These questions may ultimately converge. But they are not the same question, and a field that cannot distinguish between them cannot investigate any of them rigorously.

A post-behaviorist epistemology would begin by restoring these distinctions. It would treat ``System X achieves 90\% on benchmark Y'' as a claim about intelligence in the thinnest sense, the system can do something, perhaps even very well. It would treat ``System X reasons'' as a separate claim requiring process-level evidence. And it would treat ``System X understands'' as a further claim requiring evidence about the system's relationship to the content it processes, evidence that may or may not be obtainable with current methods, but that should not be ruled out before the question has been asked.

\subsection{Construct-Thinking as an Alternative}

We propose the term \emph{construct-thinking} for the epistemological alternative to what we have been calling the Turing framework. Where Turing-thinking asks ``does the behavior convince?'', construct-thinking asks ``does the evidence warrant the attribution?'' Where Turing-thinking treats behavioral sufficiency as an axiom, construct-thinking treats it as an empirical question: in any given case, was the behavior actually produced by a process that warrants the cognitive claim being made, or could it have been produced by a process that doesn't? Where Turing-thinking collapses the distinctions between intelligence, thinking, and understanding, construct-thinking preserves them and demands separate evidence for each.

Construct-thinking draws on the psychometric concept of construct validity \citep{cronbach1955,messick1989}, the requirement that a measurement instrument be shown to measure the theoretical construct it claims to measure, through a specified evidential procedure. Under construct-thinking, the claim that a system possesses a cognitive property (reasoning, understanding, intelligence) is treated as a construct claim: a claim about a latent property that must be supported by converging evidence from multiple sources, including but not limited to behavioral performance.

The shift from Turing-thinking to construct-thinking does not require resolving the philosophical questions about machine consciousness or understanding that Turing sought to avoid. It requires something more modest: the recognition that the field is already making claims about cognitive properties (every time a benchmark is labeled as measuring ``reasoning'' or ``understanding''), and that these claims deserve the same evidential scrutiny that construct claims receive in every other field that makes them.

\subsection{The Research Agenda}

A construct-thinking epistemology would reorient the field's research agenda. Benchmarks would need to specify their target constructs with enough precision to generate falsifiable predictions, and the jingle fallacy, assuming that because a benchmark is labeled ``reasoning,'' high performance constitutes evidence of reasoning, would be recognized as an error rather than accepted as a convention. Intelligence attributions would need to be supported by evidence beyond behavioral performance: perturbation robustness, transfer structure, mechanistic analysis, developmental trajectories, failure mode analysis. The standard of evidence would shift from ``the output is correct'' to ``the output is correct, and the process that produced it warrants the attribution.'' For any observed performance, the field would need to articulate and test competing hypotheses about the mechanisms that could have produced it, treating the underdetermination of process by output as a scientific problem to be addressed rather than an inconvenience to be ignored. And construct claims would need to specify the conditions under which they would be falsified. If no perturbation, no transfer failure, and no mechanistic finding could count as evidence against the claim that a system ``reasons,'' then the claim is not a scientific claim. It is a definitional stipulation dressed as an empirical finding.

\section{Objections}

Several objections to this argument deserve consideration.

\paragraph{``Turing was right to sidestep the definitional problem.''} This is partly true. The definitional problem is genuine, and Turing's pragmatism was productive for engineering. But the engineering successes have now generated scientific questions---about the nature of the capabilities these systems possess---that the pragmatic framework cannot answer. A tool that was appropriate for one phase of a field's development has become a constraint in the next phase. Psychology's behaviorism was also productive for establishing the field's scientific credentials. It became a constraint when the field needed to explain, not just predict, behavior.

\paragraph{``We don't need to know \emph{how} a system works, only \emph{that} it works.''} This objection is valid for engineering but not for science. If the question is whether a system can perform a task reliably in the conditions it was tested on, behavioral evaluation suffices. But even from a purely engineering perspective, the moment you deploy a system in conditions it hasn't been tested on, or expect it to generalize to circumstances you haven't foreseen, you need to know something about the process behind the performance, because a system that arrived at correct answers through genuine reasoning will generalize differently from one that arrived at correct answers through pattern-matching. If the question is whether the system possesses a cognitive property, whether it reasons, understands, or is intelligent, then the mechanism matters all the more, because the same behavior can be produced by processes that do and do not instantiate the property. The field is making the second kind of claim on the basis of evidence that supports only the first.

\paragraph{``Mechanistic interpretability is not mature enough to replace behavioral evaluation.''} We agree. Our argument is not that behavioral evaluation should be replaced but that it should be supplemented. And the immaturity of mechanistic interpretability is not a reason to ignore process-level evidence. It is a reason to invest in developing it. Psychology's cognitive revolution did not wait for neuroimaging to mature before positing internal representations. It developed behavioral paradigms (reaction time studies, priming experiments, error analysis) that provided indirect evidence about process. AI evaluation can do the same.

\paragraph{``The cognitive revolution analogy is strained---AI systems are fundamentally different from human minds.''} They are. But the analogy is not between the objects of study. It is between the epistemological frameworks governing the study. Both behaviorist psychology and Turing-framework AI evaluation commit to the same restriction (only behavioral evidence counts) and face the same consequence (questions about process become unaskable). The analogy holds at the methodological level regardless of the ontological differences between the objects.

\section{Conclusion}

In 1950, Alan Turing set aside the question of whether machines really think in order to focus on whether they can behave as if they do. For seventy-five years, this approach has served as the epistemological foundation of the evaluation of our AI endeavours. Benchmarks test behavior. Scores measure performance. Intelligence is attributed on the basis of outputs.

This paper has argued that the simplification has become a trap. The field of artificial intelligence now builds systems whose behavioral performance spans virtually every cognitive domain. It tests them on benchmarks that label this performance with human cognitive terms like reasoning, understanding, and intelligence. And it draws inferences from benchmark scores to cognitive properties that the scores cannot, in the framework's own terms, support. The behavioral framework provides no method for asking whether a system that produces correct outputs does so through a process that warrants intelligence attribution, and yet intelligence attribution is exactly what the field, in practice, does.

The parallel to psychology's behaviorist period was structural: a productive epistemological restriction that became, over time, a barrier to the questions the field needed to ask. The resolution is not the rejection of behavioral evidence, but the rejection of its sufficiency. A post-behaviorist epistemology for AI, what we have called construct-thinking, would preserve behavioral evaluation as a necessary component of assessment while demanding supplementary evidence about process, mechanism, and internal organization.

This transition will not happen by building better benchmarks within the current framework. It will happen by changing the framework---by changing what the field treats as sufficient evidence for its most consequential claims. The tools for this change are emerging. Mechanistic interpretability, perturbation analysis, transfer studies, and developmental probing all provide windows into process that behavioral evaluation alone cannot open. What is missing is not the technology but the epistemology: the collective recognition that what a system does is not the same as what a system is, and that the distance between them is a scientific question the field has not yet learned to ask.

\bibliographystyle{plainnat}
\bibliography{beyond_behavior}

\end{document}